\documentclass{article}
\usepackage{ijcai20}

\usepackage{times}

\usepackage{soul}
\usepackage{url}
\usepackage[hidelinks]{hyperref}
\usepackage[utf8]{inputenc}
\usepackage[small]{caption}
\usepackage{graphicx}
\usepackage{amsmath}
\usepackage{booktabs}
\usepackage{textcomp}
\usepackage{multirow}
\usepackage{amssymb}
\usepackage{longtable}
\usepackage{tabularx}
\usepackage{pdflscape}
\usepackage{microtype}
\usepackage[nolist,printonlyused]{acronym}
\usepackage[boxed]{algorithm2e}
\usepackage{booktabs}
\usepackage[square,sort,comma,numbers]{natbib}
\usepackage{rotating}
\usepackage{makecell}
\usepackage[flushleft]{threeparttable}

\usepackage{color}

\renewcommand{\citep}[1]{\citeauthor{#1}~\cite{#1}}

\begin{acronym}
	\acro{ADMM}{Alternating Direction Method of Multipliers}
	\acro{AI}{Artificial Intelligence}
	\acro{ALU}{Arithmetic Logic Unit}
	\acro{API}{Application Interface}
	\acro{APoZ}{average percentage of zeros}
	\acro{APG}{Accelerated Proximal Gradient}

	\acro{CNN}{Convolutional Neural Network}
	\acro{CPD}{Canonical Polyadic Decomposition}
	
	\acro{DNN}{Deep Neural Network}
	\acro{DL}{Deep Learning}
	\acro{DSP}{Digital Signal Processor}
	\acro{DRAM}{dynamic random-access memory}
	
	\acro{FoM}{Figure of Merit}
	\acro{FLOP}{floating-point operation}
	\acro{FLOPS}{floating-point operations per second}
	\acro{FSP}{flow of solution procedure}
	\acro{FFT}{Fast Fourier Transform}
	
	\acro{GAN}{Generative Adversarial Network}
	\acro{GPU}{Graphics Processing Unit}

	\acro{HPC}{High Performance Computing}
	
	\acro{IKSS}{intra-kernel stride sparsity}
	\acro{IR}{Intermediate Representation}
	\acro{IO}{input/output}
	\acro{IPS}{instructions per second}

	\acro{KD}{Knowledge Distillation}

	\acro{MAC}{multiplier-accumulate}
	\acro{ML}{Machine Learning}
	\acro{NAS}{Neural Architecture Search}
	\acro{NLP}{Natural Language Processing}
	\acro{NLR}{No Local Reuse}
	
	\acro{OS}{Output Stationary}
	\acro{PE}{processing engine}

	\acro{QoS}{Quality of Service}

	\acro{RDM}{Representational Distance Matrix}
	\acro{RL}{Reinforcement Learning}
	\acro{RS}{Row Stationary}

	\acro{SGD}{Stochastic Gradient Descend}
	\acro{SIMD}{Single instruction, multiple data}
	\acro{SIMT}{Single instruction, multiple threads}
	\acro{STE}{Straight-Through Estimator}
	\acro{SVD}{Singular Value Decomposition}
	\acro{SoC}{System-on-Chip}

	\acro{WS}{Weight Stationary}

\end{acronym}

\newcommand{\Loss}{\mathcal{L}}

\title{Pruning Algorithms to Accelerate \aclp{CNN} for Edge Applications: A Survey}

\author{
Jiayi Liu$^1$\and
Samarth Tripathi$^1$\and
Unmesh Kurup$^1$\and
Mohak Shah$^1$\\
\affiliations
$^1$LG Electronics
  5150 Great America Pkwy
  Santa Clara
  CA
  95054
 \\
\emails
\{Jason.Liu, Samarth.Tripathi, Unmesh.Kurup, Mohak.Shah\}@lge.com
}

\begin{document}

\maketitle

\begin{abstract}
With the general trend of increasing \ac{CNN} model sizes, model compression and acceleration techniques have become critical for the deployment of these models on edge devices. 
In this paper, we provide a comprehensive survey on Pruning, a major compression strategy that removes non-critical or redundant neurons from a \ac{CNN} model.
The survey covers the overarching motivation for pruning, different strategies and criteria, their advantages and drawbacks, along with a compilation of major pruning techniques.
We conclude the survey with a discussion on alternatives to pruning and current challenges for the model compression community.
\end{abstract}

\section{Introduction}\label{sec:intro}

Deep Learning has become the de-facto approach in many \ac{ML} problems, such as computer vision, natural language processing, and robotics.  
\ac{CNN} architectures and models have surpassed human performance in many such challenges. 
These advancements are a result of innovation in various research directions, including network architectures, optimization methods, and  software frameworks.
However, these breakthroughs have come at the cost of ever increasing model sizes and computation loads. Therefore, model compression becomes an important topic when \ac{CNN} models are applied in practice, especially for edge applications. 

In \ac{ML} deployment scenarios, a lightweight compressed model has numerous advantages.
On the server side, a smaller model reduces bandwidth usage and power consumption within a data center leading to savings in operational cost.  
Further, deploying these \ac{CNN} models on the client side (embedded or edge device) comes with the concomitant advantages of privacy, low latency, and better customization \cite{mtr2018ondevice}.  
However, in such scenarios they face more restricted resource requirements and need to be carefully tuned for optimal performance. 
Hence, model compression has garnered more research interests in the recent years with advances in techniques such as Pruning, Quantization, and Low Rank Decomposition.

Since \acp{CNN} are commonly over-parameterized, pruning non-critical or redundant neurons is a reasonable option to reduce the model size and \acp{FLOP} at runtime~\cite{denil2013predicting}. 
Directly searching for the best combination of neurons to be pruned is an NP-hard problem and typically not feasible for a \ac{CNN} with millions of parameters~\cite{guo2016dynamic}.
Also, a pruned network with high sparsity may not lead to practical benefits.
Therefore, a successful pruning algorithm needs to be efficient while reducing model size, improving inference speed, and maintaining accuracy. 



In this paper, we provide a comprehensive survey of the algorithmic aspects of model pruning for \acp{CNN} with a focus on 
edge deployment.
We identify the development trends and point out the current areas of focus.  
More importantly, we identify the drawbacks and challenges of these approaches and provide users with a better understanding of the trade-offs and avenues of further study. 

\section{Pruning Methodology}\label{sec:pruning}

The problem of pruning is formulated as follows: given a labelled dataset with $N$ samples, $(x_i,y_i), i\in{1\dots N}$, find the best light-weight \ac{CNN} model that takes an input $x_i$ and predicts its label $t_i = f(w, x_i)$, where $w$ represents the model parameters.  For convolution layers, a weight $w_{c',c,i,j}$ is the 4D kernel to convert $c$ input channels into $c'$ output channels with spatial convolution over $i,j$ directions.
The prediction performance is defined as the accuracy $\sum_{i=1}^{N} \mathbb{I}_{y_i=t_i}$ and the un-pruned model minimizes the loss function (for accuracy) $\sum_{i=1}^{N}\mathcal{L}(y_i,t_i)$, served as a baseline model.

The hardware limits for edge lie in Processor architecture and speed, Memory, Power/Energy consumption, and Inference latency.  In practice, analytical proxies, such as \ac{FLOP} or number of parameters, for theoretical and computational efficacy are commonly used.  In this paper, we differentiate between the actual and analytical proxy results when necessary.

Current research in pruning is divided into two components: identifying the most promising neurons to be pruned; and training and finetuning the pruned model to recover the base model's prediction performance. A successful pruning algorithm is an iterative progression of these components as illustrated in \autoref{alg:prune_workflow}~\cite{lecun1990optimal}, and improvements in the state-of-the-art come from advances in one or both of these aspects.  Therefore, we categorize existing algorithms into these two categories for clarity in the following sections. We report their compression performance in \autoref{tab:pruning_summary_table}.

\begin{algorithm}
    \SetAlgoLined
    \caption{Workflow for model pruning.}\label{alg:prune_workflow}
    \KwData{\ac{CNN} model, training data}
    \While{Compression requirement not met {\bf or} exceed budget}
      {train model (to convergence)\; 
      compute pruning criteria\;
      prune parameters below threshold\;
      }
\end{algorithm} 

\subsection{Pruning Criteria}\label{sec:pruning_criteria}

Different heuristic criteria were developed to identify the promising structures to be pruned without harming the prediction performance.
We classify these criteria into two categories: data-agnostic and data-driven where data-agnostic techniques compute saliency criteria without using the training data directly. Early works \cite{lecun1990optimal,hassibi1993second} relied on the second-order Hessian matrix, $H_{ij} = \frac{\partial^2 \Loss}{\partial{w_i}\partial{w_j}}$, to identify the weights to be
removed without harming model predictability. However, these approaches also require intensive computation of the second-order derivatives of the weights (and matrix inversion \cite{hassibi1993second}).

To alleviate the training burden, \cite{srinivas2015data} proposed to merge weights by their value similarity. They
demonstrated their success on fully-connected layers, which may dominate the model size.

Contradictory to pruning unimportant weights all at once \cite{lecun1990optimal}, \cite{han2015learning} used simple weight values as saliency to prune them iteratively.  At each pruning iteration, the weights with L2 norms below a
given threshold are located and masked out in the subsequent training and inference stages. Despite its simplicity, the iterative procedure helps the pruned model to recover and maintain its accuracy and has been commonly used in pruning approaches.

\begin{table*} [h!]
  \centering
  \caption{Saliency measurements used in pruning.}
  \label{tab:saliency}
  \begin{tabular}{lp{2.5cm}p{6cm}p{7cm}} 
     \toprule
     & Paper       & Basic idea   & Saliency Expression       \\
     \midrule                                     
    \parbox[t]{1mm}{\multirow{7}{*}{\rotatebox[origin=c]{90}{Data Agnostic}}} &
     \cite{lecun1990optimal,hassibi1993second}  & Minimize pruned deterioration & $H_{ii}w_i^2$; $w_i^2/H^{-1}_{ii}$  \\
    & \cite{han2015learning}                              & Remove weights with small values         & $|w|$ \\
    & \cite{srinivas2015data,he2019filter}           & Weight similarity and redundancy & $||w_i-w_j||; ||w-\mathbb{E}_\text{geo}[w]||$ \\
    & \cite{lebedev2016fast,wen2016learning,li2016pruning,liu2017learning}            & Structured L1/2-norm penalty      & Penalize $\sum_{c'}w^2_{c',c,i,j}$; $\sum_{c,i,j}w^2_{c',c,i,j}$; $\sum_{c'}w^2_{c',c,i,j}$;  $\sum_{c,i,j}|w_{c',c,i,j}|$ \\ 
    & \cite{zhao2019variational} & Magnitude of Batch Norm & - \\
    & \cite{he2019filter} & Remove by filter similarity & Geometric mean \\
    \hline
    \parbox[t]{1mm}{\multirow{8}{*}{\rotatebox[origin=c]{90}{Data Agnostic}}}
    &  \cite{hu2016network}            & Remove inactive neurons (\acs{APoZ})  & $\sum_{i} \mathbb{I}_{z_i=0}$  \\
    &  \cite{molchanov2016pruning}     & Remove activations with flat gradient & $|\sum_i\frac{\Loss}{\partial z_i}z_i|$ \\
    &  \cite{luo2017thinet}            & Reconstruction error on channel pruning & $||z_i - z_i^\text{pruned}||$                                                           \\
    &  \cite{luo2017entropy}           & Entropy                                 & $\sum_m P(z_i)\log P(z_i)$, where $P(z_i)$ is probability of activations in bins                                                           \\
   &   \cite{he2017channel}            & Reconstruction error and L1 norm        & -
     \\
    &  \cite{yu2018nisp}                 & Neural Importance Score               & $s_k = |w_{i+1}|^Ts_{k+1}$\\
    & \cite{golub2019full}              & Remove insensitive neurons                      & Reset weights with small
     updates to initial value \\
     \bottomrule
  \end{tabular}
\end{table*}

The above pruning approaches can significantly reduce both model size and compute, but reducing \acp{FLOP}
is not directly linked with inference speedup, especially for deep \acp{CNN}, as the weights are replaced by
sparse matrices~\cite{lebedev2016fast}.
To alleviate such problems, pruning structural components is needed. \cite{lebedev2016fast} proposed to group weights on output channels before applying the L1-norm penalty. This structured pruning leads to a smaller and more accurate model. \cite{wen2016learning} extended the above channel-based approach to additional
structure levels, i.e. filter count, filter shape, and layer width, to regularize the model size. Instead of grouping weights' L2 norms, \cite{li2016pruning} directly treated the L1-norm of filters as criterion to prune convolution layers.  \cite{zhao2019variational} used the scaling factor of the Batch Normalization layer as the saliency measure. \cite{he2019filter} proposed to prune filters close to the geometric median values, which could be represented by other filters.

Besides using model weights only, another category of approaches is to utilize the training data directly for saliency
measures, which we call \textit{data-aware} pruning. \cite{hu2016network} analyzed \ac{APoZ} of neurons as the
criterion for weight importance. \cite{he2017channel} proposed to prune model at channel level by minimizing the
reconstruction error for input data in a channel pruning strategy. \cite{luo2017entropy} proposed to use the entropy of activations to identify the channels to be removed. \cite{yu2018nisp} proposed a neuron importance score pruning (NISP) to minimize the reconstruction error in the `final response layer'.

Other more sophisticated measures have also been investigated.  \cite{molchanov2016pruning} revisited the derivatives of the loss
function that approximated the cost of dropping a feature using the first-order derivatives and pruned network structures based on grouped activations ($z_i$ in \autoref{tab:saliency}) from the feature maps.  \cite{golub2019full} proposed to use the
magnitude of gradients to prune model during training.  
  
We summarize these criteria in \autoref{tab:saliency}. Limited by computational power, early works often focused on using the data-agnostic approach on a predefined saliency to prune weights.  However, those approaches required a predefined threshold to prune neurons which make it hard to control the final compression ratio.  More recent works have shifted to use the patterns in the inference process, (e.g., \ac{APoZ}) to compress models while maintaining model accuracy.  
Also, the unstructured pruning methods that result in sparse weight matrices, have been gradually rendered outdated by structured pruning approaches that provide a more realistic performance gain (discussed later in \autoref{sec:caveats_in_pruning}).

\subsection{Pruning Procedure}\label{sec:pruning_procedure}

\begin{figure}[htbp]
    \centering
    \includegraphics[width=0.49\textwidth]{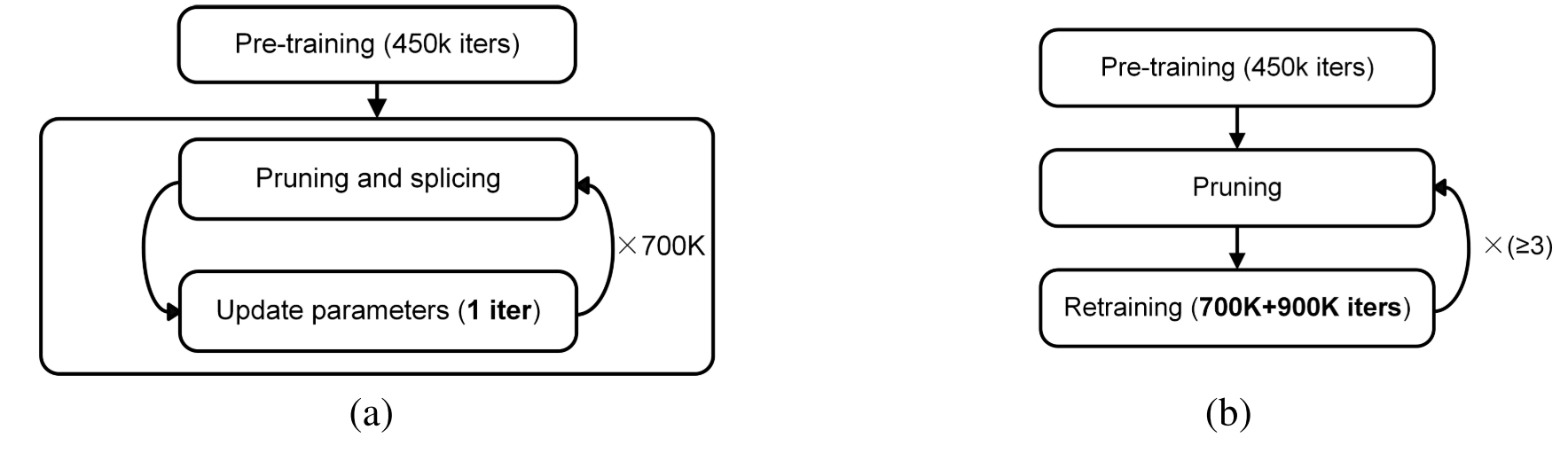} 
    \caption{Pipelines for (a) dynamic network surgery \cite{guo2016dynamic}, 
    and (b) standard pruning procedure \autoref{alg:prune_workflow}. Credit: \cite{guo2016dynamic}.}
    \label{fig:guo2016}
\end{figure}

With the above-mentioned saliency measures, early studies adopted a direct threshold on model weights and masked them
out in the following iterations. However, such pruning approaches typically deteriorate model predictive
performance and require a retraining step to recover the model prediction accuracy before further pruning (as
illustrated in \hyperref[fig:guo2016]{Fig.~\ref*{fig:guo2016}~(b)}). The iterative procedure, (training, pruning, and
then retraining) is important for a successful pruning strategy \cite{li2016pruning}. Below, we discuss
the research on improving this pruning procedure.

Selecting a pre-defined threshold has three major drawbacks as 1) the threshold is not linked to sparsity directly; 2) different layers have different sensitivity; 3) a one-time threshold may cut off too much information to restore the original accuracy. \cite{zhu2017prune} analyzed a gradual increasing threshold
schedule to prune network weights automatically to the final target sparsity. 
Another approach to solve the challenge of selecting the best thresholds for each layers, \cite{tung2017fine} adopted Bayesian optimization to automatically tune the values as a Gaussian process.  \cite{ding2018auto} proposed auto-balanced filter pruning by introducing a negative factor to L2 norm.  By using both positive and negative factors to regularize the filter weights, it can automatically select the important filters while suppressing less useful ones.

Instead of pruning with a step function, \cite{guo2016dynamic} proposed a splicing function to mask the weights.
Without an extreme change of the weight values, the pruning process can be fused with the retraining process and leads
to a significant speed up when training the compressed model, as shown in
\hyperref[fig:guo2016]{Fig.~\ref*{fig:guo2016}~(a)}). Furthermore, the soft masking helps the pruned structure to be
recovered if they are critical for the model architecture. The back-propagation is rewritten as:
\begin{equation}
\Delta w = - \eta \frac{\partial \Loss}{\partial (h(w)w)}
\end{equation}
where $h(w)$ gradually reduce unnecessary weights to zero:
\[
    h(w)= 
\begin{cases}
    0, & \text{if } a > |w|, \\
    T, & \text{if } a \leq |w| < b, \\
    1, & \text{if } b \leq |w|.
    \end{cases}
\]
Hyperparameters $a$ and $b$ control the strength of the threshold, and if $a=b$, it corresponds to the typical binary mask. One simple choice of $T$ is to use $w$ directly. 
Different from pruning by saliency, this approach can be viewed as using an auxiliary masking variable.  \cite{he2018soft} extended this idea to prune filters and demonstrated more
practical success in extensive experiments.  
\cite{liu2018frequency} applied a similar method in the frequency domain to compress \ac{CNN} models. 

Deciding which weights to drop can be considered as optimizing an objective function with a L0-norm on weights. 
Starting from this perspective, \cite{louizos2018learning} derived an equivalent expression by representing $T$ as a probabilistic gate following a Bernoulli distribution.
\cite{li2019compressing} proposed to iteratively train model and learn binary gates to prune filters,
which resulted in a compressed model with marginal decrease in accuracy. 
\cite{liu2017learning} extended the auxiliary masking variables to be learned together with model variables for filters, where the auxiliary variables scale filters and are regularized by the L1-norm. \cite{you2019gate} proposed a similar approach but with a novel tick-tock update schedule. \cite{xiao2019autoprune} treated the masking variables as scaling for the channel importance and pruned unimportant filters according to their effects on the loss.  These methods relied on auxiliary layers to learn a masking to reduce the model size efficiently.

Many L1/2 norm-based penalties mentioned earlier are the approximation to the L0-norm problem,
however they can lead to unstable and sub-optimal solutions.  Several studies worked on novel optimization methods to solve the problem. \cite{zhou2016less} translated this problem to an alternative
one and solved it by using forward-backward splitting algorithm to optimize an L2,1-norm instead. Compared to other
penalizing forms, this approach provides a better approximation to the original optimization problem.
\cite{huang2018data} adopted the \ac{APG} method to solve the problem with scaling factors similar to
\cite{guo2016dynamic} but at various structure levels to improve the performance on state-of-the-art compact
\ac{CNN} models.  \cite{li2019compressing} learned gating masks via \ac{ADMM}-based optimization. 

With advances in \ac{RL} research, there are many studies applying \ac{RL} to model pruning. \cite{huang2018learning}
built an individual gradient policy learner for each layer of a \ac{CNN} model to prune filters for an overall reward
combining the {\it accuracy} and {\it efficiency terms}.  \cite{he2018amc} adopted a
deep deterministic policy gradient to continuously control the compression ratio for a balanced target of accuracy and
resource consumption.  \cite{dong2019network} adopted \ac{NAS} to balance model performance by exploring the width and depth of a network.

In summary, the innovations for pruning procedures are classified into the following three directions of improvement: 1) efficient iterative procedure; 2) representative masking method; 3) robust learning to prune network along with training. 

\section{Discussion}

\subsection{Comparison}

\autoref{tab:pruning_summary_table} provides a compilation for model compression algorithms. We collected the results for two datasets, CIFAR-10 and ImageNet~\cite{krizhevsky2009learning,russakovsky2015imagenet}. The former is a small dataset where we expected to see significant improvement over the baseline. We reported results with VGG or ResNet models only~\cite{simonyan2014very,he2016deep}.  The ImageNet dataset is challenging but more realistic and, for that reason, we included other popular \ac{CNN} models, namely, AlexNet and MobileNet~\cite{krizhevsky2012imagenet,howard2017mobilenets}  

We reported the best pruned model with the following three measures: 
\begin{itemize}
    \item {\bf Accuracy reduction} measures the model degradation as the difference in accuracy between the pruned model and the original model. A smaller value is better and a negative value indicates that the pruned model is better than the original model. 
    \item {\bf Size reduction} measures the ratio of the reduction in size (i.e., number of parameters) over the original model size. 
    \item {\bf Time reduction} measures the ratio of the reduction in time or  \ac{FLOP} over the original model.
\end{itemize}

Overall, we observed that pruning leads to between a 10\% and 90\% reduction on both size and inference time. As commonly expected, the compression for a more challenging problem such as the ImageNet dataset, is harder than for the CIFAR-10 dataset. Unexpectedly, we found that a reduction in size is not linearly correlated with an improvement in inference speed. More profoundly, we found that only few studies have tested their performance on physical devices, which is often not well captured by \ac{FLOP}.  Finally, we found that there is limited research using the data-aware approach for filter pruning, making that an interesting direction for further study.

\begin{table*}[ht]
\caption[Summary of Pruning performance]{Summary of pruning performance. The papers are ordered by publication year for each model.}\label{tab:pruning_summary_table}

\begin{threeparttable}
\begin{tabular}{lp{0.4cm}p{1.9cm}p{1.4cm}p{1.4cm}p{1.4cm}|p{0.4cm}p{1.9cm}p{1.4cm}p{1.4cm}p{1.4cm}}
 \toprule 
 & \multicolumn{5}{c}{CIFAR-10} & \multicolumn{5}{c}{ImageNet (Top-1)}\\ \hline
 & Ref  & \thead{Model} & Accuracy \newline Reduction & Size \newline Reduction & Time \newline Reduction & Ref  & \thead{Model} & Accuracy \newline Reduction & Size \newline Reduction & Time \newline Reduction \\ \midrule 
 \parbox[t]{1mm}{\multirow{14}{*}{\rotatebox[origin=c]{90}{Data Agnostic}}} 
 & \cite{li2016pruning} & VGG-16(c)\tnote{1} & \textminus0.2\% & 64.0\% & 34.2\% & \cite{han2015learning} & AlexNet & 0.0\% & 89.0\% & 70\% \\ 
 & \cite{ding2018auto} & VGG-16(c) & 0.3\% &  & 79.7\% & \cite{lebedev2016fast}  & AlexNet(c) & 1.4\% & 69.0\% & 68.8\%(p) \\
 & \cite{huang2018learning} & VGG-16(c) & 0.6\% & 83.3\% & 48.8\%(p)\tnote{2} & \cite{wen2016learning}  & AlexNet(c) & \textminus0.1\% & - & 28.6\%(p)\\
 & \cite{li2016pruning} & ResNet-56(c) & 0.0\% & 14.1\% & 27.3\% & \cite{guo2016dynamic}  & AlexNet & \textminus0.3\% & 94.3\% & - \\
 & \cite{he2018amc} & ResNet-56 & 0.9\% & - & $\sim$ 50\% & \cite{liu2018frequency} & AlexNet & \textminus0.2\% & 95.6\% & -\\
 & \cite{he2018soft}  & ResNet-56(c) & 0.2\% & - & 52.6\% & \cite{han2015learning}  & VGG-16 & \textminus0.2\% & 92.5\% & 79\% \\ 
 & \cite{ding2018auto} & ResNet-56(c) & 0.1\% &  -  & 60.9\% & \cite{huang2018data}  & VGG-16(c) & 3.9\% & 5.6\% & 75\% \\
 & \cite{li2019compressing} & ResNet-56(c) & \textminus0.2\% & 43.1\% & 42.8\% & \cite{he2018amc}  & VGG-16 & 1.4\% & - & 80.0\% \\ 
 &&&&&&\cite{lebedev2016fast}  & VGG-19(c) & 0.7\% & 55.0\% & 50.0\%(p) \\
 &&&&&&\cite{golub2019full}  & ResNet-18 & 0.4\% & 91.5\% & -  \\
 &&&&&&\cite{huang2018data}  & ResNet-50(c) & 0.7\% & 0.8\% & 15\% \\
 &&&&&&\cite{he2018soft}  & ResNet-50(c) & 1.5\% & - &  41.8\% \\
 &&&&&& \cite{li2019compressing}  & ResNet-50(c) & 0.47\% & 42.4\% & 46.1\% \\
 &&&&&&\cite{he2018amc}  & MobileNetV1 & 0.7\% & - & 34.6\%(p) \\
 
 \hline
 
 \parbox[t]{1mm}{\multirow{7}{*}{\rotatebox[origin=c]{90}{Data Aware}}} 
 & \cite{yu2018nisp} & ResNet-56 & 0.03\% & 42.6\% & 43.6\% & \cite{yu2018nisp} & AlexNet &  0.0\% & 47.1\% & 40.1\%\\
 & \cite{you2019gate} & ResNet-56(c) & 0.03\% & 66.7\% & 70.3\% & \cite{molchanov2016pruning}  & VGG-16 & \textminus2.3\%\tnote{3}  & 34\% & 41.2\%(p) \\
 &&&&&& \cite{hu2016network}  & VGG-16 & \textminus2.5\% & 52.6\% & $\sim$ 50\% \\
 &&&&&& \cite{luo2017entropy}  & VGG-16 & \textminus0.8\% & 94.0\% & 70.0\% \\
 &&&&&& \cite{luo2017entropy}  & ResNet-50 & \textminus0.1\% & 16\% & 17.4\% \\ 
 &&&&&& \cite{yu2018nisp}  & ResNet-50 & 0.2\% & 27.1\% & 27.3\% \\
 &&&&&& \cite{you2019gate} & ResNet-50(c) & 0.7\% & 53.4\% & 55.1\% \\
 \bottomrule
\end{tabular}
\begin{tablenotes}
\item[1] The mark (c) means that the pruning is for the convolutional structure only. Occasionally, the fully connected layers have been converted to convolution layers and been pruned.
\item[2] The mark (p) means that the time reduction is measured on physical devices, otherwise, it is based on the \ac{FLOP}.
\item[3] Top-5 accuracy difference is reported.
\end{tablenotes}
\end{threeparttable}
\end{table*}

\subsection{Other Approaches}

Model compression techniques have varying advantages and disadvantages when compared with each other. Low Rank Decomposition \cite{denil2013predicting} uses linear algebra to reduce the model's weight matrices with rank decomposition. This approach allows for a mathematically sound reduction in model size and computation speed up. However, training each model requires a custom and complicated implementation procedure, which is a challenge. Quantization \cite{sze2017efficient} is another popular compression technique which involves replacing high precision floating points in \ac{CNN} weight matrices with lower precision representations. It has the distinct advantage of being both universal across different models and offering consistent performance improvements (depending on the implementation). However, it requires innate hardware support for these gains to be realized. Moreover, precision sensitivity can lead to performance deterioration. Finally, handcrafted smaller architectures such as MobileNet~\cite{howard2017mobilenets} have also enjoyed success in deployment under edge scenarios, however designing a custom architecture for different edge settings results in excessive cost and effort, and cannot be scaled.

Another interesting direction is to dynamically prune the network at runtime. \cite{dong2017more} and \cite{hua2019channel} used additional structure 
to filter the unimportant features and reduced the runtime for inference. With small auxiliary connections,
\cite{gao2019dynamic} boost the important features and suppress the irrelevant ones to reduce computation and improve inference speed. 

\subsection{Combined Approaches}

We also observed few studies on combining different compression approaches. \cite{yu2017compressing} combined low rank decomposition with pruning by threshold and successfully compressed the model size by more than 10 times for AlexNet and VGG-16 models, without losing accuracy on the ImageNet dataset.
\cite{ullrich2017soft} used the minimum description length principle to achieve quantization and pruning coherently via Bayesian variational inference.  However, an ablation study on the effectiveness of joint approaches is still missing and a systematic study would be extremely helpful to guide practitioners in this field.

\subsection{Advantages and Limitations}\label{sec:caveats_in_pruning}

\ac{CNN} pruning has gained attention alongside the rapid development in \ac{CNN} research.  It provides the following benefits:

\begin{itemize}
    \item {\bf Model size reduction}: leveraging on the redundant information carried in a \ac{CNN} model, a natural consequence of pruning is the reduced size of an input model, making it an important step for deploying models on the edge.

    \item {\bf Inference time reduction}: most of the pruning methods targeted to \ac{CNN} models provide structural pruning rather than pruning at the individual weight level.  These algorithms lead to a realistic inference time reduction for deployment.

    \item {\bf Universal compression approach}:  pruning methods are generally model independent, which can be applied to any given model architecture with minor modification.  Meanwhile, a pruned model can be deployed to current hardware environment without extensive engineering for implementation, as compared to the quantization methods.
\end{itemize}

There are still a few limitations prevent pruning becoming practical in industry settings for edge deployment.

\begin{itemize}
    \item {\bf Long training time}: the typical workflow of network pruning, i.e. \autoref{alg:prune_workflow}, requires
    iterative model training for each pruned network, which can significantly increase the time required to build these models.
    Approaches such as \cite{guo2016dynamic} speed up the training and pruning with a better update schema, but there is
    still no one-shot solution to prune a network with minimal cost;
    \item {\bf Extensive hyperparameter finetuning}: all of the pruning strategies require a set of hyperparameters to
    finely balance the compression ratio and the model accuracy. Detailed analysis of the weights on each layer was
    required in the earlier approaches while recent techniques replace such requirements with more general global parameters
    and rely on hyperparameter optimization techniques (e.g. \cite{liu2019auptimizer}) to speed up the tuning process;
    \item {\bf Small but not fast}: as realized in many later studies, a plain pruning strategy might result in fewer
    weights and \acp{FLOP}, but it does not guarantee a faster model with less energy consumption.  Building a realistic
    model to capture the correct physical resource consumption is a critical step to achieving practical model
    compression at runtime \cite{cai2018proxylessnas,marculescu2018hardware}.
    \item {\bf Benchmark}: \ac{CNN} architectures have experienced rapid development in recent years.
    Therefore, early pruning results, which are based on the over-parameterized \ac{CNN} models, such as AlexNet and VGG-16 \cite{krizhevsky2012imagenet,simonyan2014very}, may not be effective for current efficient models.  More recent studies have gradually focused on light-weight networks, such as ResNet and MobileNet \cite{he2016deep,howard2017mobilenets}.  However, a commonly accepted baseline is still missing for fair comparison.
\end{itemize}

Last but not least, the improvement in energy efficiency is not included in the above analysis. Most research in pruning
 reduce the amount computation, theoretically. However, other factors such as transferring data on to and off the chip takes energy comparable to that used for computation making the applicability of these studies limited when it comes to deployment on low power devices.
To address this discrepancy, \cite{yang2017designing} included energy consumption analysis in their iterative pruning procedure.
\cite{yang2019energyconstrained} showed that a magnitude-based pruning on SqueezeNet~\cite{iandola2016squeezenet} can
lead to higher energy consumption.  Therefore, they proposed to model the energy consumption for each layer and prune
layers based on their energy consumption  to address this issue. \cite{yang2019ecc} further modeled the energy
consumption by bilinear regression, and optimized the energy using \ac{ADMM} framework with the original loss.

The consensus procedure is that taking a pre-trained model and then iteratively retraining the model to adjust to the reduced
representation leads to a better model than training the pruned model from scratch
\cite{han2015learning,li2016pruning,he2017channel}. However, \cite{liu2018rethinking} found contradicting results that
a model trained and pruned iteratively does not provide a significantly better performance than a pruned model trained 
from scratch for a given budget. This discrepancy is another area that needs further investigation.

\section{Conclusion}

As neural network models get larger and the push towards edge/IoT devices becomes more pronounced, there is need for techniques and best practices that allow for the creation of smaller efficient models. 
Pruning is one such technique that allows us to create a smaller model from an existing larger and over-parameterized model.
In this paper we examined the constraints and metrics that motivate model compression, and formulated requirements of pruning algorithms.

We organized research in the field based on pruning criteria and pruning procedure. We further classified the pruning criteria into data-agnostic and data-aware approaches.  Our hope is that this breakdown will provide guidance for future researchers to develop new algorithms and allow them to compare previous works effectively (in addition to the comprehensive comparison in \autoref{tab:pruning_summary_table}).

The field of deep learning is accelerated by the development of tools and frameworks for model building and training. Except for a few examples such as PocketFlow and Distiller~\cite{wu2018pocketflow,neta2018neural}, a general platform for pruning is generally missing. Our work also serves as a guidance to develop a universal compression pipeline by depicting the independent components and procedures in the pruning process.

Finally, we provided an in-depth discussion on the limitations, advantages and novel research directions along with a comparison of the performance of major pruning techniques on standard datasets and models. 

\bibliographystyle{plain}
{ \small
\bibliography{ijcai20}

\begin{thebibliography}{10}

\bibitem{cai2018proxylessnas}
Han Cai, Ligeng Zhu, and Song Han.
\newblock Proxyless{NAS}: Direct neural architecture search on target task and
  hardware.
\newblock In {\em International Conference on Learning Representations}, 2019.

\bibitem{denil2013predicting}
Misha Denil, Babak Shakibi, Laurent Dinh, Marc'Aurelio Ranzato, and Nando
  De~Freitas.
\newblock Predicting parameters in deep learning.
\newblock In {\em Advances in neural information processing systems}, pages
  2148--2156, 2013.

\bibitem{ding2018auto}
Xiaohan Ding, Guiguang Ding, Jungong Han, and Sheng Tang.
\newblock Auto-balanced filter pruning for efficient convolutional neural
  networks.
\newblock In {\em AAAI Conference on Artificial Intelligence}, 2018.

\bibitem{dong2017more}
Xuanyi Dong, Junshi Huang, Yi~Yang, and Shuicheng Yan.
\newblock More is less: A more complicated network with less inference
  complexity.
\newblock In {\em Proceedings of the IEEE Conference on Computer Vision and
  Pattern Recognition}, pages 5840--5848, 2017.

\bibitem{dong2019network}
Xuanyi Dong and Yi~Yang.
\newblock Network pruning via transformable architecture search.
\newblock In H.~Wallach, H.~Larochelle, A.~Beygelzimer, F.~d\textquotesingle
  Alch\'{e}-Buc, E.~Fox, and R.~Garnett, editors, {\em Advances in Neural
  Information Processing Systems 32}, pages 759--770. 2019.

\bibitem{gao2019dynamic}
Xitong Gao, Yiren Zhao, {\L}ukasz Dudziak, Robert Mullins, and Cheng zhong Xu.
\newblock Dynamic channel pruning: Feature boosting and suppression.
\newblock In {\em International Conference on Learning Representations}, 2019.

\bibitem{golub2019full}
Maximilian Golub, Guy Lemieux, and Mieszko Lis.
\newblock Full deep neural network training on a pruned weight budget.
\newblock In {\em SysML}, 2019.

\bibitem{guo2016dynamic}
Yiwen Guo, Anbang Yao, and Yurong Chen.
\newblock Dynamic network surgery for efficient dnns.
\newblock In {\em Advances In Neural Information Processing Systems}, pages
  1379--1387, 2016.

\bibitem{han2015learning}
Song Han, Jeff Pool, John Tran, and William Dally.
\newblock Learning both weights and connections for efficient neural network.
\newblock In {\em Advances in neural information processing systems}, pages
  1135--1143, 2015.

\bibitem{hassibi1993second}
Babak Hassibi and David~G Stork.
\newblock Second order derivatives for network pruning: Optimal brain surgeon.
\newblock In {\em Advances in neural information processing systems}, pages
  164--171, 1993.

\bibitem{he2016deep}
Kaiming He, Xiangyu Zhang, Shaoqing Ren, and Jian Sun.
\newblock Deep residual learning for image recognition.
\newblock In {\em Proceedings of the IEEE conference on computer vision and
  pattern recognition}, pages 770--778, 2016.

\bibitem{he2018soft}
Yang He, Guoliang Kang, Xuanyi Dong, Yanwei Fu, and Yi~Yang.
\newblock Soft filter pruning for accelerating deep convolutional neural
  networks.
\newblock In {\em Proceedings of the 27th International Joint Conference on
  Artificial Intelligence}, pages 2234--2240. AAAI Press, 2018.

\bibitem{he2019filter}
Yang He, Ping Liu, Ziwei Wang, Zhilan Hu, and Yi~Yang.
\newblock Filter pruning via geometric median for deep convolutional neural
  networks acceleration.
\newblock In {\em The IEEE Conference on Computer Vision and Pattern
  Recognition (CVPR)}, June 2019.

\bibitem{he2018amc}
Yihui He, Ji~Lin, Zhijian Liu, Hanrui Wang, Li-Jia Li, and Song Han.
\newblock Amc: Automl for model compression and acceleration on mobile devices.
\newblock In {\em Proceedings of the European Conference on Computer Vision
  (ECCV)}, pages 784--800, 2018.

\bibitem{he2017channel}
Yihui He, Xiangyu Zhang, and Jian Sun.
\newblock Channel pruning for accelerating very deep neural networks.
\newblock In {\em Proceedings of the IEEE International Conference on Computer
  Vision}, pages 1389--1397, 2017.

\bibitem{howard2017mobilenets}
Andrew~G Howard, Menglong Zhu, Bo~Chen, Dmitry Kalenichenko, Weijun Wang,
  Tobias Weyand, Marco Andreetto, and Hartwig Adam.
\newblock Mobilenets: Efficient convolutional neural networks for mobile vision
  applications.
\newblock {\em arXiv preprint arXiv:1704.04861}, 2017.

\bibitem{hu2016network}
Hengyuan Hu, Rui Peng, Yu-Wing Tai, and Chi-Keung Tang.
\newblock Network trimming: A data-driven neuron pruning approach towards
  efficient deep architectures.
\newblock {\em arXiv preprint arXiv:1607.03250}, 2016.

\bibitem{hua2019channel}
Weizhe Hua, Yuan Zhou, Christopher~M De~Sa, Zhiru Zhang, and G.~Edward Suh.
\newblock Channel gating neural networks.
\newblock In H.~Wallach, H.~Larochelle, A.~Beygelzimer, F.~d\textquotesingle
  Alch\'{e}-Buc, E.~Fox, and R.~Garnett, editors, {\em Advances in Neural
  Information Processing Systems 32}, pages 1884--1894, 2019.

\bibitem{huang2018learning}
Qiangui Huang, Kevin Zhou, Suya You, and Ulrich Neumann.
\newblock Learning to prune filters in convolutional neural networks.
\newblock In {\em 2018 IEEE Winter Conference on Applications of Computer
  Vision (WACV)}, pages 709--718. IEEE, 2018.

\bibitem{huang2018data}
Zehao Huang and Naiyan Wang.
\newblock Data-driven sparse structure selection for deep neural networks.
\newblock In {\em Proceedings of the European Conference on Computer Vision
  (ECCV)}, pages 304--320, 2018.

\bibitem{iandola2016squeezenet}
Forrest~N Iandola, Song Han, Matthew~W Moskewicz, Khalid Ashraf, William~J
  Dally, and Kurt Keutzer.
\newblock Squeezenet: Alexnet-level accuracy with 50x fewer parameters and
  \textless 0.5 mb model size.
\newblock {\em arXiv preprint arXiv:1602.07360}, 2016.

\bibitem{mtr2018ondevice}
{MIT} Technology~Review Insights.
\newblock On-device processing and ai go hand-in-hand.
\newblock
  \url{https://www.technologyreview.com/s/610421/on-device-processing-and-ai-go-hand-in-hand/},
  March 2018.
\newblock Accessed: 2019-09-11.

\bibitem{krizhevsky2009learning}
Alex Krizhevsky.
\newblock Learning multiple layers of features from tiny images.
\newblock 2009.

\bibitem{krizhevsky2012imagenet}
Alex Krizhevsky, Ilya Sutskever, and Geoffrey~E Hinton.
\newblock Imagenet classification with deep convolutional neural networks.
\newblock In {\em Advances in neural information processing systems}, pages
  1097--1105, 2012.

\bibitem{lebedev2016fast}
Vadim Lebedev and Victor Lempitsky.
\newblock Fast convnets using group-wise brain damage.
\newblock In {\em Proceedings of the IEEE Conference on Computer Vision and
  Pattern Recognition}, pages 2554--2564, 2016.

\bibitem{lecun1990optimal}
Yann LeCun, John~S Denker, and Sara~A Solla.
\newblock Optimal brain damage.
\newblock In {\em Advances in neural information processing systems}, pages
  598--605, 1990.

\bibitem{li2016pruning}
Hao Li, Asim Kadav, Igor Durdanovic, Hanan Samet, and Hans~Peter Graf.
\newblock Pruning filters for efficient convnets.
\newblock {\em arXiv preprint arXiv:1608.08710}, 2016.

\bibitem{li2019compressing}
Tuanhui Li, Baoyuan Wu, Yujiu Yang, Yanbo Fan, Yong Zhang, and Wei Liu.
\newblock Compressing convolutional neural networks via factorized
  convolutional filters.
\newblock In {\em Proceedings of the IEEE Conference on Computer Vision and
  Pattern Recognition}, pages 3977--3986, 2019.

\bibitem{liu2019auptimizer}
Jiayi Liu, Samarth Tripathi, Unmesh Kurup, and Mohak Shah.
\newblock Auptimizer -- an extensible, open-source framework for hyperparameter
  tuning.
\newblock In {\em The IEEE BigData 2019}, 2019.

\bibitem{liu2018frequency}
Zhenhua Liu, Jizheng Xu, Xiulian Peng, and Ruiqin Xiong.
\newblock Frequency-domain dynamic pruning for convolutional neural networks.
\newblock In {\em Advances in Neural Information Processing Systems}, pages
  1043--1053, 2018.

\bibitem{liu2017learning}
Zhuang Liu, Jianguo Li, Zhiqiang Shen, Gao Huang, Shoumeng Yan, and Changshui
  Zhang.
\newblock Learning efficient convolutional networks through network slimming.
\newblock In {\em Proceedings of the IEEE International Conference on Computer
  Vision}, pages 2736--2744, 2017.

\bibitem{liu2018rethinking}
Zhuang Liu, Mingjie Sun, Tinghui Zhou, Gao Huang, and Trevor Darrell.
\newblock Rethinking the value of network pruning.
\newblock {\em arXiv preprint arXiv:1810.05270}, 2018.

\bibitem{louizos2018learning}
Christos Louizos, Max Welling, and Diederik~P Kingma.
\newblock Learning sparse neural networks through $ l\_0 $ regularization.
\newblock 2018.

\bibitem{luo2017entropy}
Jian-Hao Luo and Jianxin Wu.
\newblock An entropy-based pruning method for cnn compression.
\newblock {\em arXiv preprint arXiv:1706.05791}, 2017.

\bibitem{luo2017thinet}
Jian-Hao Luo, Jianxin Wu, and Weiyao Lin.
\newblock Thinet: A filter level pruning method for deep neural network
  compression.
\newblock In {\em Proceedings of the IEEE international conference on computer
  vision}, pages 5058--5066, 2017.

\bibitem{marculescu2018hardware}
Diana Marculescu, Dimitrios Stamoulis, and Ermao Cai.
\newblock Hardware-aware machine learning: modeling and optimization.
\newblock In {\em Proceedings of the International Conference on Computer-Aided
  Design}, page 137. ACM, 2018.

\bibitem{molchanov2016pruning}
Pavlo Molchanov, Stephen Tyree, Tero Karras, Timo Aila, and Jan Kautz.
\newblock Pruning convolutional neural networks for resource efficient
  inference.
\newblock {\em arXiv preprint arXiv:1611.06440}, 2016.

\bibitem{russakovsky2015imagenet}
Olga Russakovsky, Jia Deng, Hao Su, Jonathan Krause, Sanjeev Satheesh, Sean Ma,
  Zhiheng Huang, Andrej Karpathy, Aditya Khosla, Michael Bernstein, et~al.
\newblock Imagenet large scale visual recognition challenge.
\newblock {\em International journal of computer vision}, 115(3):211--252,
  2015.

\bibitem{simonyan2014very}
Karen Simonyan and Andrew Zisserman.
\newblock Very deep convolutional networks for large-scale image recognition.
\newblock {\em arXiv preprint arXiv:1409.1556}, 2014.

\bibitem{srinivas2015data}
Suraj Srinivas and R.~Venkatesh Babu.
\newblock Data-free parameter pruning for deep neural networks.
\newblock In {\em Procedings of the British Machine Vision Conference 2015},
  pages 31.1--31.12. British Machine Vision Association, 2015.

\bibitem{sze2017efficient}
Vivienne Sze, Yu-Hsin Chen, Tien-Ju Yang, and Joel~S Emer.
\newblock Efficient processing of deep neural networks: A tutorial and survey.
\newblock {\em Proceedings of the IEEE}, 105(12):2295--2329, 2017.

\bibitem{tung2017fine}
Frederick Tung, Srikanth Muralidharan, and Greg Mori.
\newblock Fine-pruning: Joint fine-tuning and compression of a convolutional
  network with bayesian optimization.
\newblock {\em arXiv preprint arXiv:1707.09102}, 2017.

\bibitem{ullrich2017soft}
Karen Ullrich, Edward Meeds, and Max Welling.
\newblock Soft weight-sharing for neural network compression.
\newblock 2017.

\bibitem{wen2016learning}
Wei Wen, Chunpeng Wu, Yandan Wang, Yiran Chen, and Hai Li.
\newblock Learning structured sparsity in deep neural networks.
\newblock In {\em Advances in neural information processing systems}, pages
  2074--2082, 2016.

\bibitem{wu2018pocketflow}
Jiaxiang Wu, Yao Zhang, Haoli Bai, Huasong Zhong, Jinlong Hou, Wei Liu, and
  Junzhou Huang.
\newblock Pocketflow: An automated framework for compressing and accelerating
  deep neural networks.
\newblock In {\em Advances in Neural Information Processing Systems (NIPS),
  Workshop on Compact Deep Neural Networks with Industrial Applications}, 2018.

\bibitem{xiao2019autoprune}
XIA XIAO, Zigeng Wang, and Sanguthevar Rajasekaran.
\newblock Autoprune: Automatic network pruning by regularizing auxiliary
  parameters.
\newblock In H.~Wallach, H.~Larochelle, A.~Beygelzimer, F.~d\textquotesingle
  Alch\'{e}-Buc, E.~Fox, and R.~Garnett, editors, {\em Advances in Neural
  Information Processing Systems 32}, pages 13681--13691. 2019.

\bibitem{yang2019ecc}
Haichuan Yang, Yuhao Zhu, and Ji~Liu.
\newblock Ecc: Platform-independent energy-constrained deep neural network
  compression via a bilinear regression model.
\newblock In {\em Proceedings of the IEEE Conference on Computer Vision and
  Pattern Recognition}, pages 11206--11215, 2019.

\bibitem{yang2019energyconstrained}
Haichuan Yang, Yuhao Zhu, and Ji~Liu.
\newblock Energy-constrained compression for deep neural networks via weighted
  sparse projection and layer input masking.
\newblock In {\em International Conference on Learning Representations}, 2019.

\bibitem{yang2017designing}
Tien-Ju Yang, Yu-Hsin Chen, and Vivienne Sze.
\newblock Designing energy-efficient convolutional neural networks using
  energy-aware pruning.
\newblock In {\em Proceedings of the IEEE Conference on Computer Vision and
  Pattern Recognition}, pages 5687--5695, 2017.

\bibitem{you2019gate}
Zhonghui You, Kun Yan, Jinmian Ye, Meng Ma, and Ping Wang.
\newblock Gate decorator: Global filter pruning method for accelerating deep
  convolutional neural networks.
\newblock In H.~Wallach, H.~Larochelle, A.~Beygelzimer, F.~d\textquotesingle
  Alch\'{e}-Buc, E.~Fox, and R.~Garnett, editors, {\em Advances in Neural
  Information Processing Systems 32}, pages 2130--2141. 2019.

\bibitem{yu2018nisp}
Ruichi Yu, Ang Li, Chun-Fu Chen, Jui-Hsin Lai, Vlad~I Morariu, Xintong Han,
  Mingfei Gao, Ching-Yung Lin, and Larry~S Davis.
\newblock Nisp: Pruning networks using neuron importance score propagation.
\newblock In {\em Proceedings of the IEEE Conference on Computer Vision and
  Pattern Recognition}, pages 9194--9203, 2018.

\bibitem{yu2017compressing}
Xiyu Yu, Tongliang Liu, Xinchao Wang, and Dacheng Tao.
\newblock On compressing deep models by low rank and sparse decomposition.
\newblock In {\em Proceedings of the IEEE Conference on Computer Vision and
  Pattern Recognition}, pages 7370--7379, 2017.

\bibitem{zhao2019variational}
Chenglong Zhao, Bingbing Ni, Jian Zhang, Qiwei Zhao, Wenjun Zhang, and Qi~Tian.
\newblock Variational convolutional neural network pruning.
\newblock In {\em The IEEE Conference on Computer Vision and Pattern
  Recognition (CVPR)}, June 2019.

\bibitem{zhou2016less}
Hao Zhou, Jose~M Alvarez, and Fatih Porikli.
\newblock Less is more: Towards compact cnns.
\newblock In {\em European Conference on Computer Vision}, pages 662--677.
  Springer, 2016.

\bibitem{zhu2017prune}
Michael Zhu and Suyog Gupta.
\newblock To prune, or not to prune: exploring the efficacy of pruning for
  model compression.
\newblock {\em arXiv preprint arXiv:1710.01878}, 2017.

\bibitem{neta2018neural}
Neta Zmora, Guy Jacob, and Gal Novik.
\newblock Neural network distiller, June 2018.

\end{thebibliography}
}
\end{document}